\pdfoutput=1
\documentclass[11pt]{article}

\usepackage{EMNLP2022}

\usepackage{times}
\usepackage{latexsym}

\usepackage[T1]{fontenc}
\usepackage[utf8]{inputenc}

\usepackage{microtype}

\usepackage{inconsolata}

\usepackage{amsmath}
\usepackage{amssymb}
\usepackage{url}
\usepackage{booktabs} % For formal tables
\usepackage{graphicx}
\usepackage{epstopdf}
\usepackage{subfigure}
\usepackage{verbatim}
\usepackage{bm}
\usepackage{array}
\usepackage{multirow}
\usepackage{subfigure}
\usepackage{makecell}
\usepackage{xcolor}
\usepackage{xspace}
\usepackage[most]{tcolorbox}

\usepackage[font={small}]{caption}

\newcommand{\dataset}[0]{\textbf{\textsc{ConvFinQA}}\xspace}

\title{\textsc{ConvFinQA}: Exploring the Chain of Numerical Reasoning in Conversational Finance Question Answering}

\author{\textbf{Zhiyu Chen}\textsuperscript{1}, \textbf{Shiyang Li}\textsuperscript{1}, \textbf{Charese Smiley}\textsuperscript{2}, \textbf{Zhiqiang Ma}\textsuperscript{2}, \\ \textbf{Sameena Shah}\textsuperscript{2} and \textbf{William Yang Wang}\textsuperscript{1} \\
  \textsuperscript{1}University of California, Santa Barbara \\
  \textsuperscript{2}J.P. Morgan \\
  {\tt \{zhiyuchen,shiyangli,william\}@cs.ucsb.edu}, \\ {\tt \{charese.h.smiley,zhiqiang.ma,sameena.shah\}@jpmchase.com} \\}
\begin{document}
\maketitle

\begin{abstract}
With the recent advance in large pre-trained language models, researchers have achieved record performances in NLP tasks that mostly focus on language pattern matching. The community is experiencing the shift of the challenge from how to model language to the imitation of complex reasoning abilities like human beings. 
In this work, we investigate the application domain of finance that involves real-world, complex numerical reasoning. We propose a new large-scale dataset, \dataset, aiming to study the chain of numerical reasoning in conversational question answering. Our dataset poses great challenge in modeling long-range, complex numerical reasoning paths in real-world conversations. We conduct comprehensive experiments and analyses with both the neural symbolic methods and the prompting-based methods, to provide insights into the reasoning mechanisms of these two divisions. We believe our new dataset should serve as a valuable resource to push forward the exploration of real-world, complex reasoning tasks as the next research focus. 
Our dataset and code is publicly available\footnote{\url{https://github.com/czyssrs/ConvFinQA}}.

\end{abstract}
\section{Introduction}
\begin{figure}[ht]
\centering
\includegraphics[width=0.48\textwidth]{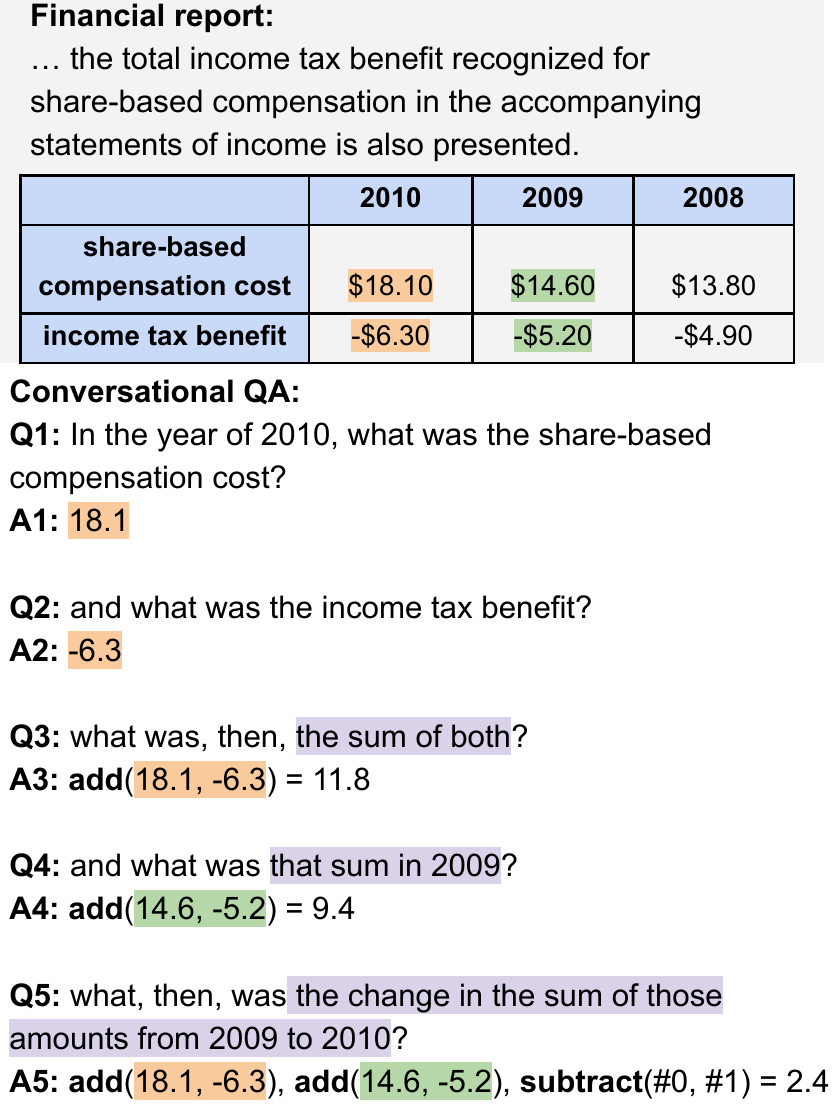}
\caption{An example from \dataset: each question may depend on previous questions to answer. } 
\label{fig:intro-eg0}
\end{figure}

The rapid advancement in developing large pre-trained language models (LM) has brought the natural language processing research into a new era. Based on the well-known transformer~\cite{DBLP:conf/nips/VaswaniSPUJGKP17} architecture, such large pre-trained LMs~\cite{DBLP:conf/naacl/DevlinCLT19,radford2019language,DBLP:journals/jmlr/RaffelSRLNMZLL20,DBLP:journals/corr/abs-2110-08207,DBLP:journals/corr/abs-2205-10747} have set up new state-of-the-art results for many NLP tasks, with some of them approaching or even surpassing human performances, like on the SQuAD~\cite{DBLP:conf/emnlp/RajpurkarZLL16} dataset. We observe that the tasks with the essence of modeling language patterns can be well addressed by large pre-trained LMs. However, for the other kind of tasks requiring complex reasoning abilities, current researches are still away from satisfactory performances~\cite{DBLP:journals/corr/abs-2201-11903}. 
% This is in accordance with the intuition that the human reasoning ability may root heterologous with language ability.  

Traditional methods on reasoning tasks typically take neural symbolic models to encode the context, generate the reasoning program and do the execution~\cite{DBLP:conf/acl/LiangBLFL17,DBLP:conf/iclr/ChenLYZSL20}. Most recently, it is shown that sufficiently large pre-trained LMs can excel at reasoning tasks given proper prompts~\cite{DBLP:journals/corr/abs-2201-11903}. But their tasks being experimented with are relatively general and toy, such as simple math word problems. The form of the solutions and the reasoning explanations probably have been witnessed by the model during pre-training. This raises an interesting question: Which of the two directions is the fundamental way to solve complex reasoning problems? 

In this work, we go beyond the simple reasoning tasks and dive into the real application domain of finance to investigate the complex numerical reasoning ability of current modeling paradigms. The finance domain bears the natural requirements of realistic, complex numerical reasoning from human labor, such as quantitative analysis of financial reports. We seek to study the real-world scenario of \textbf{conversational question answering over financial reports} -- investors or analysts would typically ask sequential questions to get insights into the numerical in the reports. The questions require extensive calculations and meanwhile often demonstrate cross dependency, forming the chains of numerical reasoning throughout the conversation. 

To this end, we propose a new dataset, \dataset (\underline{\textbf{Con}}versational \underline{\textbf{Fin}}ance \underline{\textbf{Q}}uestion \underline{\textbf{A}}nswering), with 3,892 conversations consisting 14,115 questions. To construct the dataset, we design a framework to simulate the conversation flow by decomposition and concatenation of the multi-hop questions from the FinQA~\cite{DBLP:conf/emnlp/ChenCSSBLMBHRW21} dataset. We then ask expert annotators to compose the question for each conversation turn based on the simulated conversing flow. Figure~\ref{fig:intro-eg0} shows one example conversation from our dataset. We conduct comprehensive experiments and analyses on our dataset using both the neural symbolic models and the prompting-based methods, and summarize the following insights:
(1) Both kinds of approaches (with the execution accuracy less than 70.0\%) fall far behind human performance (89.4\%). The reasoning chains throughout the conversation pose great challenges for the models to learn when to refer to or discard the conversation history and how to assemble the reasoning path. 
(2) Though excelling at simple general reasoning tasks, prompting-based methods perform a lot worse for our task (less than 50.0\% using GPT-3 175B). They either superficially mimic the given prompts or recall their own knowledge for simple general numerical reasoning. They tend to fail to understand new complex task paradigms for new domains. We believe our new dataset should serve as a challenging and valuable resource for the exploration of real-world, complex reasoning tasks as the next research focus.

\begin{table}[t]
\small
\begin{center}
\resizebox{0.45\textwidth}{!}{%
\begin{tabular}{lcccc}
\toprule
Dataset & Size & Mode & Challenge & Domain\\
\midrule
SQA & 6k & ConvQA & table navigation & general\\
CSQA & 200k & ConvQA & KG reasoning & general\\
CoQA & 8k & ConvQA & co-reference & general\\
QuAC & 14k & ConvQA & open-ended & general\\
\midrule
DROP & 96k & QA & numerical reasoning & general\\
MathQA & 37k & QA & numerical reasoning & math\\
FinQA & 8k & QA & numerical reasoning & finance\\
TAT-QA & 17k & QA & numerical reasoning & finance\\
\midrule
\dataset & 4k & ConvQA & numerical reasoning & finance\\
\bottomrule
\end{tabular}
}
\caption{Comparison of \dataset with existing datasets. }
\label{compare_data}
\end{center}
\end{table}
\section{Related Work}
\paragraph{Conversational Question Answering}
Conversational question answering (ConvQA) ~\cite{DBLP:journals/corr/abs-2106-00874} has been gaining attentions in recent years. In ConvQA, the users can append multiple questions in addition to the first one to get more information. This also mitigates the need to ask a single complex multi-hop question at one time, making the information-seeking procedure more natural. For previous datasets, SQA~\cite{DBLP:conf/acl/IyyerYC17} are built by decomposing multi-hop questions based on Wikitables. CSQA~\cite{DBLP:conf/aaai/SahaPKSC18} questions require simple logical operations over knowledge graphs (KGs). CoQA~\cite{DBLP:journals/tacl/ReddyCM19} focuses on co-references among the conversation turns to be more human-like. QuAC~\cite{DBLP:conf/emnlp/ChoiHIYYCLZ18} focuses on open-ended, exploratory questions. In contrast, our dataset \dataset targets complex numerical reasoning chains among the sequential questions in finance conversations. 

\paragraph{Numerical Reasoning} 
The numerical reasoning ability is often investigated in the form of question answering. The DROP dataset~\cite{DBLP:conf/naacl/DuaWDSS019} explores simple calculations over texts in the general domain. MaWPS~\cite{DBLP:conf/naacl/Koncel-Kedziorski16} and MathQA~\cite{DBLP:conf/naacl/AminiGLKCH19} focus on generating solutions for math word problems. Recently, \citet{DBLP:journals/corr/abs-2201-11903} demonstrate that large pre-trained LMs can excel at reasoning tasks given proper prompts with natural language explanations. However, their reasoning tasks are mostly simple and general. In this work, we explore complex numerical reasoning in a highly specialized domain. 

\paragraph{Financial NLP} 
Previous work in financial NLP mostly centers on sentiment analysis~\cite{DBLP:conf/asunam/DayL16, DBLP:conf/emnlp/AkhtarKGEB17}, fraud detection~\cite{DBLP:conf/acl/HanBHDBW18, DBLP:conf/emnlp/WangZLZL19, DBLP:journals/corr/abs-1908-09156},
opinionated QA~\cite{DBLP:conf/ijcai/0001HH0Z20}, such as the FiQA\footnote{https://sites.google.com/view/fiqa/home} dataset built based on social media. 
% Recent works attempt to develop pre-trained models specialized for finance domain~\cite{DBLP:journals/corr/abs-2006-08097, DBLP:journals/corr/abs-1908-10063}, and the downstream tasks are mostly sentiment classifications. 
% To the best of our knowledge, there is no previous work and dataset on building QA systems of numerical reasoning on financial reports. 
Most recently, \citet{DBLP:conf/emnlp/ChenCSSBLMBHRW21} propose the FinQA dataset with multi-hop numerical reasoning questions based on financial report. TAT-QA~\cite{DBLP:conf/acl/ZhuLHWZLFC20} is another QA dataset with a similar focus. In \dataset, we seek to construct question sequences in the conversational setting aiming at more natural experiences for real-world usages. Table~\ref{compare_data} presents the comparison of our dataset with existing ones. 
\section{Task Formulation}
Given a financial report containing both the textual content $T$ and structured table $B$, the user asks a sequence of questions $\{Q_i\}^{n}_{i=0}$ where later questions may depend on previous questions to answer. The target is to generate the reasoning program $G$ to be executed to get the answer $A$ to the last question:
\begin{equation}
    P(A | T, B, Q_n) = \sum P(G_i | T, B, Q_0,Q_1,...Q_{n-1})
\end{equation}
Where $\{G_i\}$ is all the possible programs to evaluate to the correct answer. We follow the same domain specific language (DSL) in FinQA~\cite{DBLP:conf/emnlp/ChenCSSBLMBHRW21} to construct the reasoning programs as a sequence of operation-argument clauses (Appendix A for all operations):
\begin{equation}
    \text{op}_{1} [\boldsymbol{\text{args}_{1}}] , \text{op}_2 [\boldsymbol{\text{args}_{2}}] ... , \text{op}_n [\boldsymbol{\text{args}_{n}}]
\end{equation}
We follow the same evaluation metric as in FinQA, the execution accuracy to evaluate the final execution result and program accuracy to evaluate program equivalence. 
\section{The \dataset Dataset}
\subsection{Dataset Construction}
\label{data_construction}
\paragraph{The Overview}
The core challenge of building such a dataset is the construction of a natural, realistic conversational flow -- what kinds of questions the queriers may ask and how these questions logically appear in a conversation. 
We consult financial experts to summarize the following key factors integrating a conversation when querying financial reports:
{\em (i)} The questioner directly queries the surface content.
{\em (ii)} The questioner asks something requiring calculations from the numbers in the report to answer.
{\em (iii)} The questioner asks the above two kinds of questions sequentially to form the conversation, to cumulatively query more information or switch to other aspects.
\begin{figure}[h]
\centering
\includegraphics[width=0.48\textwidth]{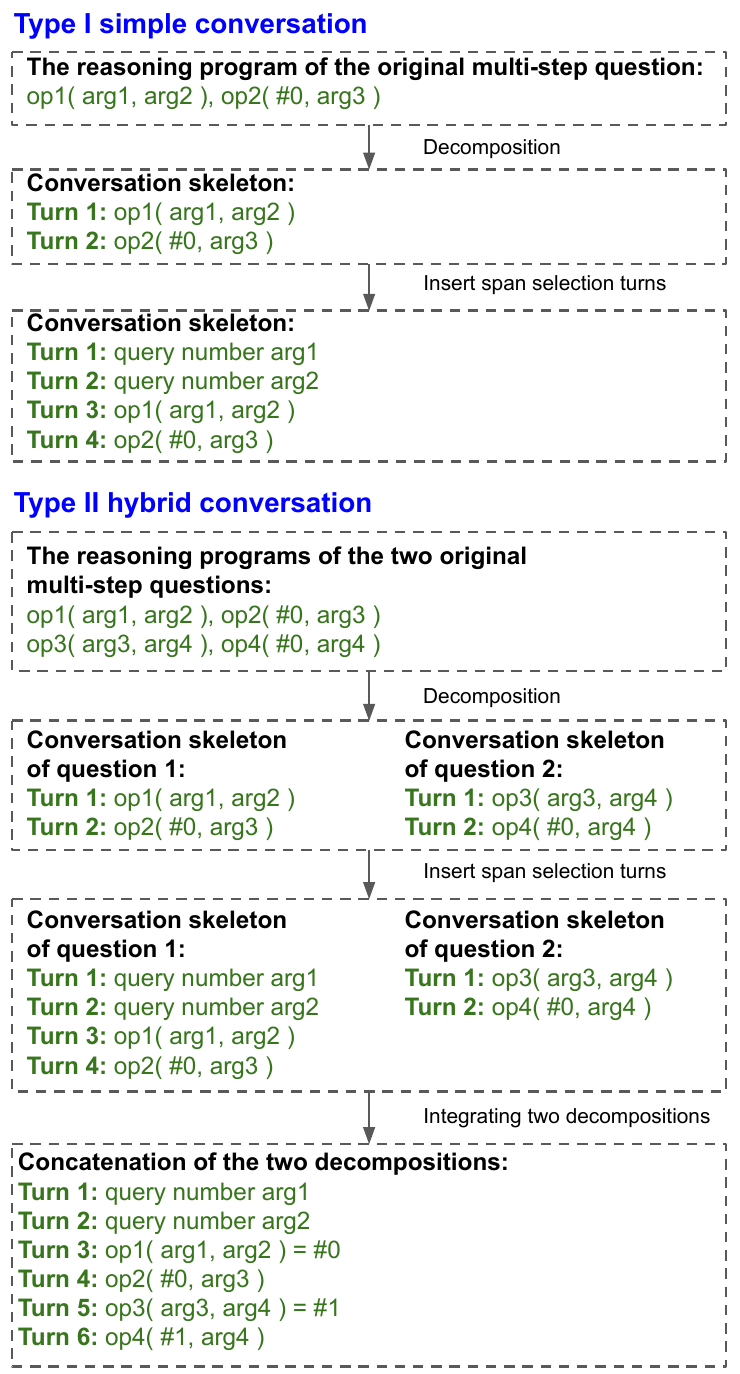}
\caption{The simulation process of conversation skeletons.} 
\label{fig:data_simulate}
\end{figure}

Directly composing the conversations from scratch involving all the above factors is very heavy and costly. 
To tackle this challenge, we propose a two-step construction framework: \textbf{(I): Conversational QA flow simulation} to produce the conversation skeleton with each turn filled with the reasoning semantics, and \textbf{(II): Question composition} to realize the reasoning semantics into textual questions. 
% Figure~\ref{fig:data_construct} gives an overview of our dataset construction framework. 
\begin{figure*}[ht]
\centering
\includegraphics[width=1.0\textwidth]{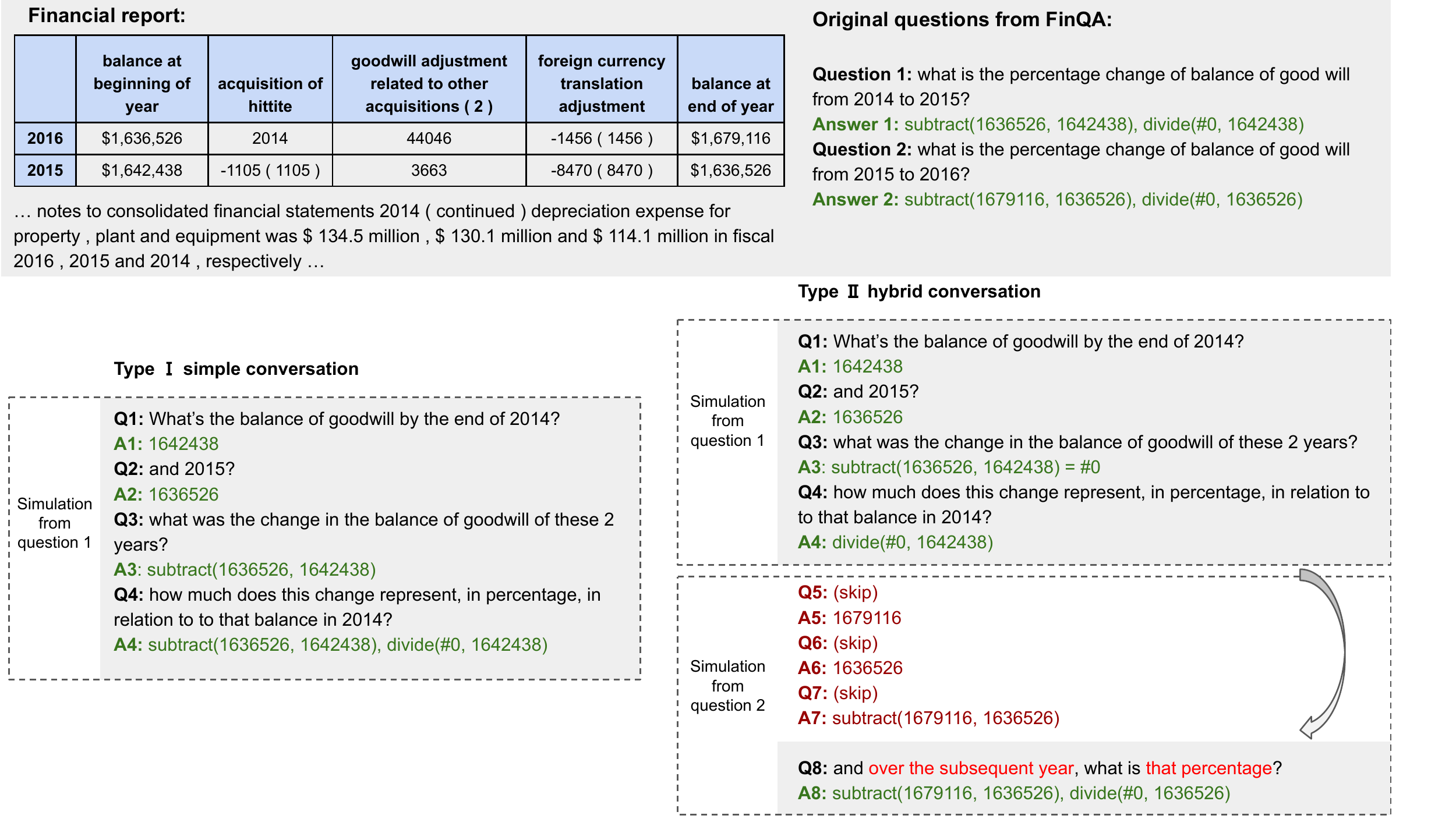}
\caption{The question composition examples for the two types of conversations. For the hybrid conversation example, the annotator skips three turns and directly jumps to the last turn using references, making the conversation more natural. } 
\label{fig:data_construct}
\end{figure*}
\paragraph{Conversational QA Flow Simulation}
We build the conversation flow based on the decomposition and concatenation of the multi-step reasoning programs (the solutions of the multi-hop questions) in the existing FinQA~\cite{DBLP:conf/emnlp/ChenCSSBLMBHRW21} dataset. In FinQA, the authors construct two multi-hop questions for most of its reports. The two FinQA questions for the same report naturally query different but sometimes correlated aspects of the report, inspiring us to integrate them into a natural and realistic conversation. We simulate two types of conversations: \textbf{Type I: Simple conversation} from the decomposition of a single multi-hop question and \textbf{Type II: Hybrid conversation} from the decomposition and integration of two multi-hop questions. Figure~\ref{fig:data_simulate} illustrates the simulation processes of the two types of conversation flows. 

For \textbf{Type I simple conversations}, we take one muti-hop question and decompose its reasoning program into single steps -- each reasoning step will then be realized into one question as one conversation turn. 
To consider the scenario that the questioner directly queries the surface content, every time there is a new number in a reasoning step, we randomly insert an additional turn before this turn with the semantic to query this new number.

For \textbf{Type II hybrid conversations}, we take two multi-hop questions based on the same report, decompose their reasoning programs and insert additional number selection turns similar to the type I conversation. Then we concatenate the decompositions of the two questions to integrate the full conversation skeleton -- corresponding to the scenario where the questioner asks two different aspects of the same report. Since the two aspects of the same report often correlate with each other, the conversation flow constructed this way will involve longer dependencies among the turns. 

\paragraph{Question Composition}
After we construct both types of conversation skeletons, we employ expert annotators to realize the skeletons into textual questions. 
We use the UpWork\footnote{\url{www.upwork.com}} platform to recruit expert annotators with finance backgrounds, such as CPAs, MBAs, etc. Figure~\ref{fig:data_construct} gives the composition examples of the two types of conversations. 

Specifically, we present the financial report and the simulated conversation skeletons to the annotators, with each turn filled with the reasoning semantics (the decomposed reasoning program or a single number for the number selection turn). We instruct the annotators to: 
{\em (i)} Read the report and understand the reasoning flow of the whole conversation skeleton; 
{\em (ii)} Compose questions for the turns based on the given reasoning semantics; 

Since our conversation skeletons are simulated, there must be many unnatural scenarios, e.g., unnatural decompositions, redundant or unnecessary turns, etc. Therefore, we emphasize the following key points:
{\em (i)} The annotators can skip some turns and directly jump to a certain following turn with the goal of an overall natural conversation. The key is to \textit{identify redundancies} in the given conversation flow and compress the unnecessary turns using references to the previous context. The right example in Figure~\ref{fig:data_construct} shows a scenario to skip unnecessary turns. 
{\em (ii)} If there is no way to compose a natural conversation with the given skeleton, the annotators can discard the example. 
We launch training sessions for the annotators to master task settings before working on the official large batches.

\subsection{Dataset Analysis}
\label{data_analysis}
\paragraph{Dataset Statistics}
We end up with 3,892 conversations containing 14,115 questions. We split the dataset into 3,037/421/434 for train/dev/test sets. 2,715 of the conversations are simple conversations, and the rest 1,177 are hybrid conversations. Table~\ref{table:gen_stats} summarizes the general statistics of our dataset. 

\begin{figure}[ht]
\centering
\includegraphics[width=0.48\textwidth]{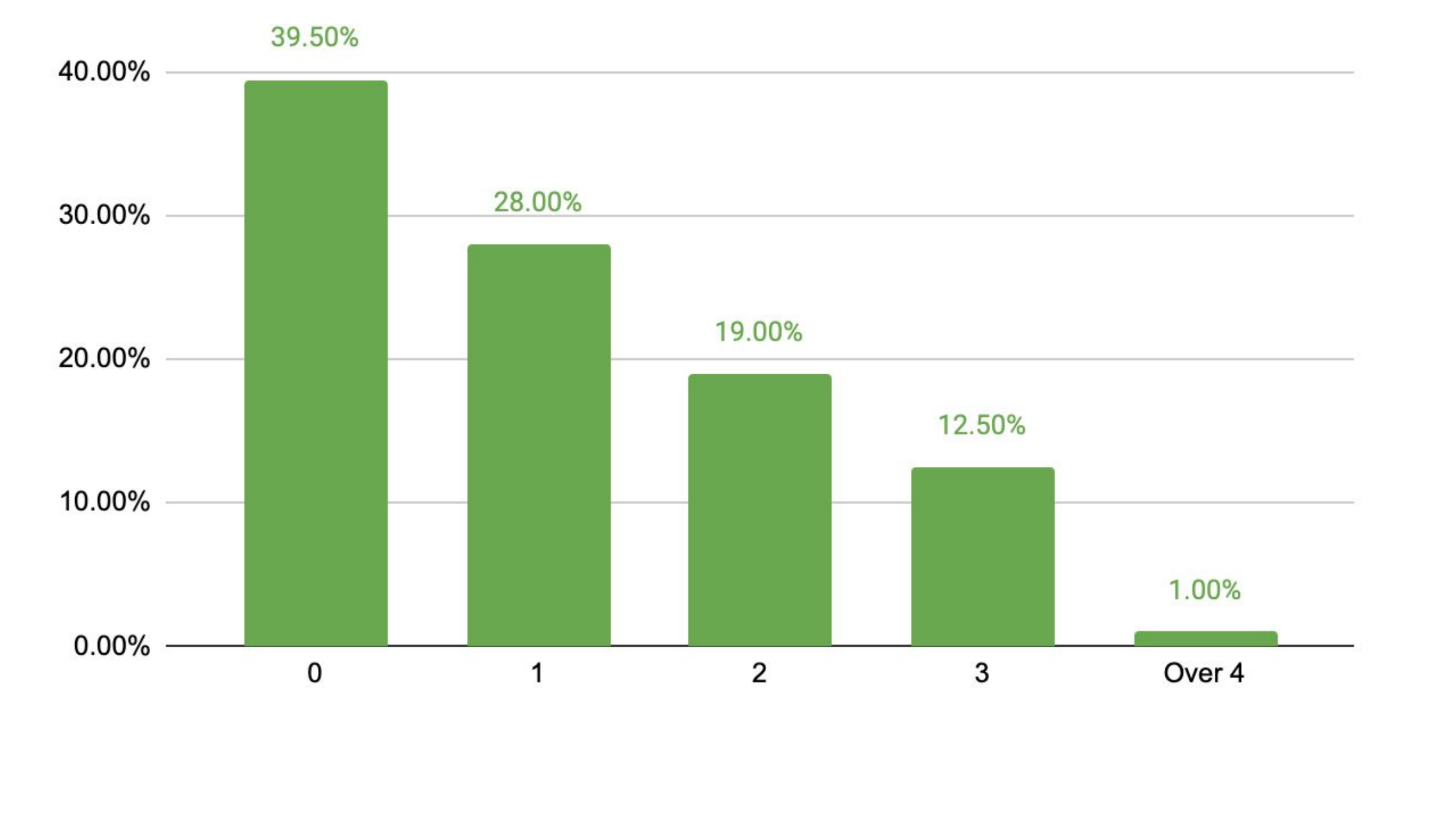}
\caption{Distribution of the longest dependency distances of the questions in \dataset. Over 60\% of the questions have longer dependencies with previous questions. } 
\label{fig:dependency}
\end{figure}

In our \dataset dataset, the major challenge is to learn the chain of numerical reasoning throughout the conversation turns. First, we sample 200 turns from our dataset and ask the expert annotators to count the longest dependency distance to answer the current question, i.e., how many previous questions need to be seen to answer the current one. Figure~\ref{fig:dependency} shows the result distributions. Second, in \dataset, we build two types of conversations -- the simple conversation from the decomposition of one FinQA question and the hybrid conversation from the decompositions and concatenation of two FinQA questions. We are interested to see, for the second type of hybrid conversations, how the question set from the second FinQA question makes references to the first one. We split the hybrid conversations into the two sets -- one from the first source FinQA question and one from the second, and ask the expert annotators to decide whether any questions from the second set depend on the questions from the first set to answer. Among 200 samples, 65.0\% of them depend on the first question set to answer, which demonstrates the challenging reasoning chains in our dataset -- the model may need to construct the reasoning chains crossing different aspects and long-range. At last, we also classify the type of questions based on the reasoning forms of the answers. 34.73\% of the questions are number selection questions, 35.10\%, 25.41\%, and 4.75\% of them have reasoning programs of 1, 2, and over 3 steps, respectively.

For the type of questions, 59.18\% of the questions rely on supporting facts only from the table to answer, 25.56\% of the questions rely on supporting facts only from the text, and the rest 15.26\% rely on both. 
For the types of calculations, we have around 18.80\% additions, 40.49\% subtractions, 6.92\% multiplications, 33.43\% divisions.

\begin{table}[t]
\small
\begin{center}
\resizebox{0.45\textwidth}{!}{%
\begin{tabular}{lr}
\toprule

Conversations & 3,892\\
Questions & 14,115\\
Report pages & 2,066\\
Vocabulary & 20k\\
Avg. \# questions in one conversation & 3.67\\
Avg. question length & 10.59\\
Avg. \# sentences in input text & 23.65\\
% Avg. \# tokens in input text & 615.08\\
Avg. \# rows in input table & 6.39\\
% Avg. \# tokens in input table & 60.53\\
Avg. \# tokens in all inputs (text \& table) & 675.61\\
Max. \# tokens in all inputs (text \& table) & 2338.00\\
\bottomrule
\end{tabular}
}
\caption{Statistics of \dataset.}
\label{table:gen_stats}
\end{center}
\end{table}

% \begin{figure}[ht]
% \centering
% \includegraphics[width=0.48\textwidth]{figs/stat_steps.pdf}
% \caption{Question type distributions of \dataset: } 
% \label{fig:stat_steps}
% \end{figure}

\paragraph{Data Quality Assessment}
To evaluate the quality of \dataset and establish human performance references, we sample 200 example questions and distribute to both the expert and laymen annotators to answer. The two expert annotators reach an average execution accuracy of 89.44\% and program accuracy of 86.34\%, with an agreement rate over 85.0\% for both metrics. For the laymen performance, we distribute the samples to MTurk\footnote{Three built-in worker qualifications are used: HIT Approval Rate ($\geq $95\%), Number of Approved HITs ($\geq 1000$), and Locale (US Only) Qualification. We do not select any professional constraints. We pay \$2.0 for each question.} and end up with an execution accuracy of 46.90\% and program accuracy of 45.52\% with agreement rates lower than 60.0\%. This again demonstrates the great expertise required to solve our dataset. 
\section{Experiments on Neural Symbolic Approaches}
\label{nsa_all}
In this section, we will first experiment with traditional neural symbolic approaches using the full training data and make detailed analyses. 

\subsection{Methods and Main Results}
\label{nsa_main_res}
We take the FinQANet model from~\cite{DBLP:conf/emnlp/ChenCSSBLMBHRW21} and two generative models -- the GPT-2~\cite{radford2019language} and T5~\cite{DBLP:journals/jmlr/RaffelSRLNMZLL20}. FinQANet is a pipeline approach with a retriever to first retrieve the supporting facts from the financial report, then a generator taking the supporting facts and the question as the input to decode the reasoning program. Structural information and constraints are also involved in the decoder. 
We adopt the same retrieving process from FinQANet and use the current conversation context, i.e., the questions up to the current turn, to retrieve the evidences from the input financial report. We end up with the retrieval results of 86.38\% recall for the top 3 retrieved facts. For the program generation, we concatenate the retrieved facts with the conversation context as the input. We experiment with the encoder varied as BERT~\cite{DBLP:conf/naacl/DevlinCLT19} and RoBERTa~\cite{DBLP:journals/corr/abs-1907-11692}. 
Table~\ref{table:main_res} shows the overall experiment results on \dataset. Using a specially designed encoder-decoder with structural preservation of the program, FinQANet still outperforms the standalone generative models. While there is still a gap till the expert performance, the models already surpass the laymen performance. We can see that such neural approaches specially designed can learn better numerical reasoning ability for the specific domain than the \textit{common sense numerical reasoning ability} of the general crowd. 

\begin{table}[t]
\small
\begin{center}
\resizebox{.45\textwidth}{!}{%
\begin{tabular}{lcc}
\toprule
\textbf{Baselines} & \textbf{Exe Acc} & \textbf{Prog Acc}\\
\midrule
GPT-2(medium) & 58.19 & 57.00 \\
\midrule
T-5(large) & 58.66 & 57.05 \\
\midrule
FinQANet (BERT-base) & 55.03 & 54.57 \\
\midrule
FinQANet (BERT-large) & 61.14 & 60.55 \\
\midrule
FinQANet (RoBERTa-base) & 64.95 & 64.16 \\
\midrule
FinQANet (RoBERTa-large) & \textbf{68.90} & \textbf{68.24} \\
\midrule
\midrule
FinQANet-Gold (RoBERTa-large) & 77.32 & 76.46 \\
\midrule
\midrule
Human Expert Performance & 89.44 & 86.34 \\
\midrule
General Crowd Performance & 46.90 & 45.52 \\
\bottomrule
\end{tabular}
}
%\caption{Our main results on test set.}
\caption{The execution accuracy (Exe Acc) and program accuracy (Prog Acc) for the models. We also experiment with using gold supporting facts, shown as FinQANet-Gold. }
\label{table:main_res}
\end{center}
\end{table}

\subsection{Performance Breakdown}
\begin{figure}[ht]
\centering
\includegraphics[width=0.45\textwidth]{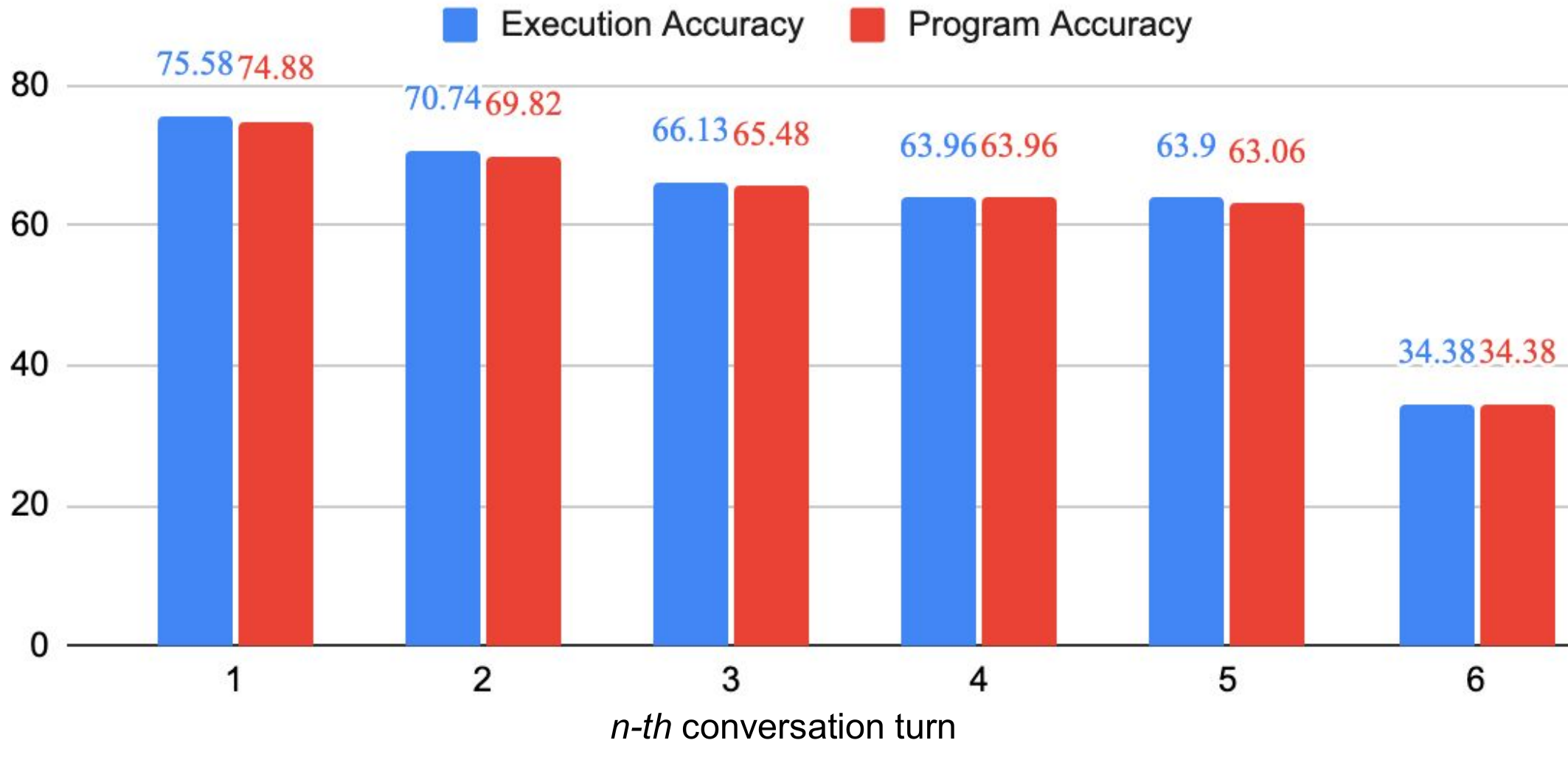}
\caption{Performances for the \textit{nth} conversation turn.} 
\label{fig:turn_stat}
\end{figure}

To gain a deeper understanding of the model insights, we analyze the performances of different types of questions. The results are shown in Table~\ref{table:detail_res}. We can see that the number selection turns are the easiest to answer. Considering different types of conversations, the hybrid conversations are harder to learn than simple conversations, especially the second part of the hybrid conversations where the question set comes from the decomposition of the second multi-hop question. In these questions, some of them are irrelevant to the questions in the first part, while some of them depend on the questions from the first part to answer. The model faces a stronger challenge of finding the correct reasoning chains. We also look into the performance breakdown by conversation turns, which is shown in Figure~\ref{fig:turn_stat}. Later turns in the conversations tend to be harder to answer due to longer reasoning dependencies.

\begin{table}[t]
\small
\begin{center}
\resizebox{.45\textwidth}{!}{
\begin{tabular}{lcc}
\toprule
\textbf{Methods} & \textbf{Exe Acc} & \textbf{Prog Acc}\\
\midrule
\textbf{full results} & 68.90 & 68.24 \\
\midrule
\multicolumn{2}{l}{\textbf{Number selection turns}} & \\
\midrule
Number selection questions & 82.54 & 82.34 \\
\midrule
Program questions & 62.14 & 61.26 \\
\midrule
\multicolumn{2}{l}{\textbf{Simple \& hybrid conversations}} & \\
\midrule
Simple conversations & 72.37 & 72.00 \\
\midrule
Hybrid conversations & 60.99 & 59.70 \\
\midrule
Hybrid conversations (first part) & 68.11 & 66.54 \\
\midrule
Hybrid conversations (second part) & 52.38 & 51.43 \\

\bottomrule
\end{tabular}
}
\caption{Performance breakdown. The number selection questions are the easiest to answer. The hybrid conversations are harder than simple conversations, while the second part of them is even more difficult.}
\label{table:detail_res}
\end{center}
\end{table}

\subsection{Analyses and Findings}
\begin{figure*}[ht]
\centering
\includegraphics[width=0.96\textwidth]{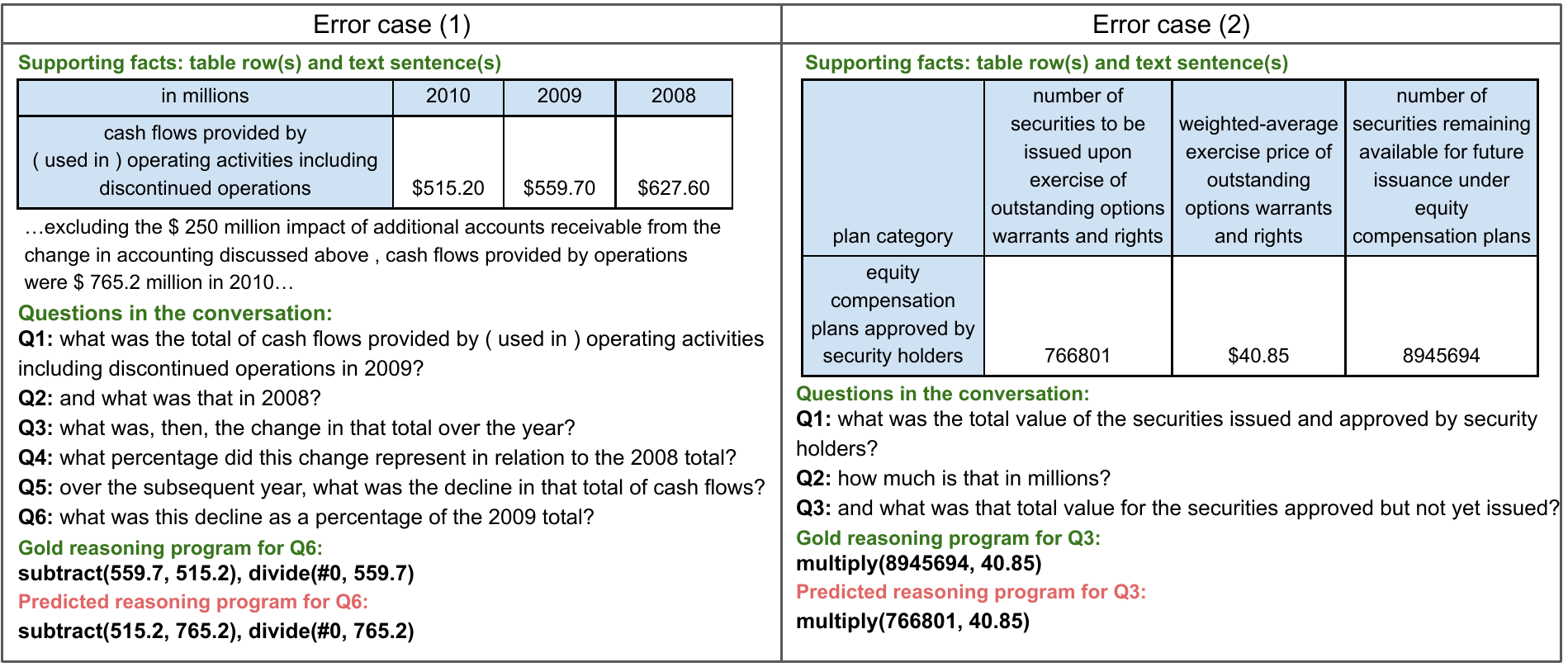}
\caption{Error cases from the results of FinQANet(RoBERTa-large)} 
\label{fig:case_1}
\end{figure*}

We manually analyze a sample of the predictions from the FinQANet(RoBERTa-large) model and summarize the following findings: 
\paragraph{The model excels at number selection questions. } For the number selection questions depending on previous references, e.g., \textit{what is that value in the subsequent year?}, the model is mostly able to answer. Also, the model is mostly clear on when to discard the previous context and make the transition to new questions. 
\paragraph{The model suffers from the lack of domain knowledge. } The lack of financial knowledge leads to many errors of missing retrieval facts, wrong value selections, and wrong mathematical generations. Nonetheless, the current large pre-trained models do see financial corpus during pre-training; we still need to endow the system with stronger domain knowledge for tasks requiring high-level, complex domain reasoning abilities. 
\paragraph{The model struggles with long reasoning chains. } For the later question turns in a conversation that demonstrate longer reasoning dependencies to the previous context, the model often struggles with deducting the correct reasoning programs. If the prediction for any turn is wrong, then there is a very minor chance that the subsequent turns are correct. 
We provide two error case studies in Figure~\ref{fig:case_1}. 

\section{Experiments on Prompting-Based Approaches}
In this section we attempt on few-shot learning with prompting-based methods and reveal the insights. 

\subsection{Methods and Main Results}
We use the GPT-3 text-davinci-002 model\footnote{OpenAI has released the model interface as a paid service}~\cite{DBLP:conf/nips/BrownMRSKDNSSAA20}. 
Directly injecting the full financial report into the prompt is not realistic because of the length constraint. Therefore we still attempt the retriever-generator paradigm. 
Due to the high cost of using GPT-3, in this work we only run retrieval on a sample of the test set, and run program generation on the full test set using the gold retrieval results as the input. Nonetheless, we believe our experiments are sufficient to show many interesting and valuable insights into the prompting-based methods on \dataset. 

For the retrieval, we concatenate each sentence or linearized table row of the report with the conversation context, and let the model predict if the former is relevant for answering the last question. We use 16 exemplars and run GPT-3 on a sample of 300 examples of the test set. We end up with an average recall of 74.25\% using 3 different sets of exemplars, which is much lower than the retriever trained with the full training data in \S\ref{nsa_main_res}. 

For the program generation, the exemplar is formatted as \texttt{[supporting facts, conversation context, result to be generated]}. We experiment using the following settings: \textbf{(i) Answer-only}, to directly generate the execution results. \textbf{(ii) Program-original}, to generate the reasoning program with the original DSL. \textbf{(iii) Program-normal}, to generate the reasoning program with the normal DSL. We convert the programs into the normal form commonly used in the general domain, e.g., \texttt{$add(a_1, a_2) \rightarrow a_1 + a_2$}.
\textbf{(iv) the Chain of Thought (CoT) prompting.} CoT prompting~\cite{DBLP:journals/corr/abs-2201-11903} includes a natural language explanation of the reasoning steps before reaching the answer. We ask 3 expert annotators to compose the explanations for the exemplars.
For each method, we run experiments using 3 sets of 10 different exemplars. Table~\ref{table:gpt_main_res} shows the overall results. Even with the gold retrieval results, GPT-3 still underperforms the neural symbolic approaches with full-training data in \S\ref{nsa_all}. See Appendix D for all the prompt details.

\begin{table}[t]
\small
\begin{center}
\resizebox{.4\textwidth}{!}{%
\begin{tabular}{lcc}
\toprule
\textbf{Baselines} & \textbf{Exe Acc} & \textbf{Prog Acc}\\
\midrule
Answer-only & $24.09_{0.61}$ & - \\
\midrule
Program-original & $40.81_{4.68}$ & $36.62_{4.22}$ \\
\midrule
Program-normal & $45.15_{2.77}$ & $38.88_{2.57}$ \\
\midrule
CoT prompting & $40.63_{1.25}$ & $33.84_{2.19}$ \\
\midrule
Human Expert Performance & 89.44 & 86.34 \\
\midrule
General Crowd Performance & 46.90 & 45.52 \\
\bottomrule
\end{tabular}
}
%\caption{Our main results on test set.}
\caption{The results for all the prompting methods. We report the average and the standard deviation of different sets of exemplars or annotators. }
\label{table:gpt_main_res}
\end{center}
\end{table}

\subsection{Performance Breakdown}
\label{gpt_breakdown}
We take the results from the best-performing method, \textbf{Program-normal}, to investigate the detailed performances. Table~\ref{table:gpt_detail_res} shows the performance breakdown for different types of turns. Surprisingly, GPT-3 even performs worse on number selection turns. We find that the model often makes errors for the number selection turns with references to the previous conversation context, e.g., for the question \textit{what is that value in the subsequent year?}, the model still chooses the value in the previous year. Even if we specify the conversational QA setting in the prompt instructions and explicitly ask to answer the last question, the model likely does not understand this task paradigm and often fails to make correct references to the context. This further makes the performances in longer reasoning chains worse, as shown in Table~\ref{table:gpt_detail_res}. We also analyze the performances for conversation turn length and exemplar numbers in Appendix D. 

\begin{table}[t]
\small
\begin{center}
\resizebox{.45\textwidth}{!}{
\begin{tabular}{lcc}
\toprule
\textbf{Methods} & \textbf{Exe Acc} & \textbf{Prog Acc}\\
\midrule
\textbf{full results} & 48.85 & 42.14 \\
\midrule
\multicolumn{2}{l}{\textbf{Number selection turns}} & \\
\midrule
Number selection questions & 35.32 & 34.72 \\
\midrule
Program questions & 55.56 & 45.82 \\
\midrule
\multicolumn{2}{l}{\textbf{Simple \& hybrid conversations}} & \\
\midrule
Simple conversations & 52.22 & 46.64 \\
\midrule
Hybrid conversations & 41.16 & 31.90 \\
\midrule
Hybrid conversations (first part) & 56.30 & 48.03 \\
\midrule
Hybrid conversations (second part) & 22.85 & 12.38 \\

\bottomrule
\end{tabular}
}
\caption{Performance breakdown: it is hard for GPT-3 to learn the problem paradigm and correctly make references to the conversation context. Still, questions with longer context is harder to answer.}
\label{table:gpt_detail_res}
\end{center}
\end{table}

% \begin{figure}[ht]
% \centering
% \includegraphics[width=0.48\textwidth]{figs/turn_stats_gpt.pdf}
% \caption{Performances for question of the \textit{nth} conversation turn.} 
% \label{fig:turn_stat_gpt}
% \end{figure}

\subsection{Analyses and Findings}
We analyze samples of the predictions for all the methods and summarize the following findings:
\paragraph{GPT-3 can do simple calculations by itself. } For methods that generate the reasoning programs, compared with the results of the neural symbolic approaches in \S\ref{nsa_all}, the gap between the execution and program accuracy is much larger. We find that GPT-3 often directly generates the correct numerical results without generating the program. Though the given prompts always derive the programs first, GPT-3 tends to use its own knowledge acquired during pre-training. This is also the reason why Answer-only achieves certain correctness. However, GPT-3 still struggles with complex calculations, such as long digits and divisions.

\paragraph{GPT-3 performs better for its familiar program format. }In Table~\ref{table:gpt_main_res}, Program-normal outperforms Program-original, since we use the common form of calculation which is seen much more frequently by GPT-3 during its pre-training. GPT-3 makes many grammar errors for Program-original. 

\paragraph{GPT-3 struggles with new complex task paradigms. }Like stated in \S\ref{gpt_breakdown}, GPT-3 probably has not seen a similar paradigm as our task setting during pre-training. We see many examples where GPT-3 simply mimics the reasoning steps given in one exemplar but ignores the actual context. This is also the reason that CoT prompting performs even worse than generating the program only. We explicitly explain our task setting in the prompt instructions about how the questions in the conversation are interrelated and the task goal to answer the current turn. However, in many cases, GPT-3 either mimics the reasoning steps given in the exemplars or comes up with incorrect reasoning based on its own knowledge in the general domain.
See Appendix D for error cases from Program-normal.

\section{Conclusion and Discussion}
Our new dataset, \dataset, targets one of the major directions to be explored as the next research focus -- how to simulate human reasoning abilities in complex real-world settings. We experiment with the neural symbolic models with full training data and the prompting-based few-shot learning and find that:
(1) Both approaches are still away from human expert performances, indicating the challenge of this task. 
(2) The neural symbolic approach uses specifically crafted architectures to learn co-occurrence patterns with large-scale training data. The prompting-based approach recalls its own memory of elaborating the reasoning process with the trigger of the prompts. However, this may not work well when encountering new complex task paradigms for new domains. 
(3) Theoretically, we may encode as many task paradigms into the large LMs, as long as the reasoning process can be clearly illustrated by language. But for highly specialized domains or tasks, designing specific models also tend to be more realistic and effective.
(4) We are also eager to see the actual \textit{bound} between the reasoning tasks that can benefit from language modeling and the ones that can not. This should be the crucial factor in deciding the upper bound of what large LMs can solve with reasoning. 
\section{Limitations}
In this work, we investigate two construction mechanisms for the conversation, the decomposition of single multi-hop questions and the decomposition and concatenation of two multi-hop questions regarding the same report. This definitely does not cover all possible cases in real-world conversations. We make this first attempt and hope for future work to continue exploration. 

For prompting-based methods, we only experiment with the GPT-3 model, whose interface is released to the public as a paid service. Also, due to cost constraints, we do not conduct extensive experiments on complex prompt engineering. We believe our experiments can provide valuable insights into the task of complex reasoning over real-world specific domains, and meanwhile we do not exclude the possibility that there could be better performances for prompting-based methods if applying advanced prompt engineering or even larger pre-trained LMs, like the PaLM model~\cite{DBLP:journals/corr/abs-2204-02311} which is not released. We leave this for future work. 
\section{Ethical Considerations}

\paragraph{Dataset Collection Process and Conditions.}
For the annotation of our \dataset dataset on Upwork, we first launch interviews of the task introduction with 2 example conversations, which is paid as \$30. Then based on their consents to continue working on the large-scale job, we discuss with the workers to reach agreements on the compensation before starting the large-scale job. For the simple conversations from one FinQA question, we pay around \$4.0 per conversation. For complex conversations from two FinQA questions, we pay around \$7.0 per conversation. The hourly rates are discussed and agreed upon with both sides based on the working speed of different workers. Among all the US-based hires, the average hourly rate is \$60.0, with the minimum hourly rate of \$50.0. The evaluation tasks and prompt writing tasks follow the similar procedure and rates. 

\paragraph{IRB (Institutional Review Board) Approval.}
The dataset annotation is classified as exempt by our Institutional Review Board (IRB). The systems trained using our dataset are primarily intended to be used as augmenting human decision-making in financial analysis, but not as a replacement of human experts.

\section*{Acknowledgment}
We thank the anonymous reviewers for their thoughtful comments. This research was supported by the J.P. Morgan Faculty research award. The authors are solely responsible for the contents of the paper and the opinions expressed in this publication do not reflect those of the funding agencies.

\bibliography{anthology,custom}
\bibliographystyle{acl_natbib}

\appendix

\section*{Appendix A: Operation Definitions}

We describe all the operations in Table~\ref{table:def_op}.

\begin{table*}[ht]
\centering
\resizebox{0.6\textwidth}{!}{%
\begin{tabular}{l|l|l|l}
\toprule
Name & Arguments & Output & Description \\
\midrule
add & number1, number2 & number & add two numbers: $number1 + number2$ \\
\midrule
subtract & number1, number2 & number & subtract two numbers: $number1 - number2$ \\
\midrule
multiply & number1, number2 & number & multiply two numbers: $number1 \cdot number2$ \\
\midrule
divide & number1, number2 & number & multiply two numbers: $number1 / number2$ \\
\midrule
exp & number1, number2 & number & exponential: $number1^{number2}$ \\
\midrule
greater & number1, number2 & bool & comparison: $number1 > number2$ \\
\bottomrule
\end{tabular}
}
\caption{Definitions of all operations}
\label{table:def_op}
\end{table*}

\section*{Appendix B: Annotation Interface}
\begin{figure*}[ht]
\centering
\includegraphics[width=\textwidth]{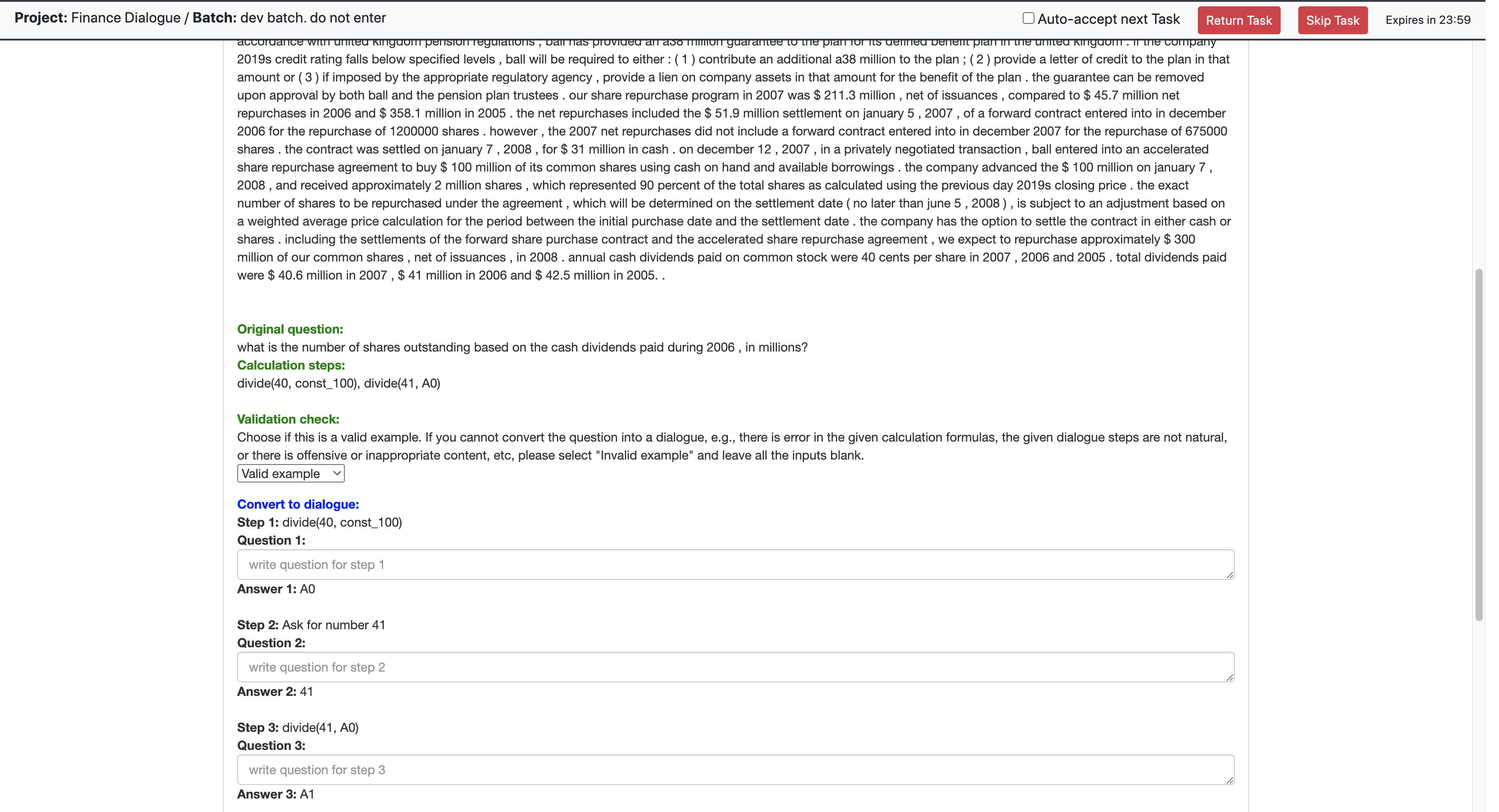}
\caption{Annotation interface. } 
\label{fig:ui_0}
\end{figure*}

We use Turkle\footnote{https://github.com/hltcoe/turkle} to build our annotation platform, which is a Django-based web application that can run in a local server. Figure~\ref{fig:ui_0} shows our annotation interface. We present the financial report and the decomposed program list to the annotators and ask them to re-write each program step into a question.

\section*{Appendix C: Experiment Details}
For the neural symbolic approaches, the training of all models are conducted on TITAN RTX GPUs. All the implementation and pre-trained models are based on the huggingface transformers library. We use the Adam optimizer~\cite{DBLP:journals/corr/KingmaB14}. The learning rate of all models varies at the level of 1e-5 (except for T-5 with 1e-4). We set the batch size as 16. Table~\ref{table:valid_res} shows the results on validation set. 

\begin{table}[t]
\small
\begin{center}
\resizebox{.4\textwidth}{!}{%
\begin{tabular}{lcc}
\toprule
\textbf{Baselines} & \textbf{Exe Acc} & \textbf{Prog Acc}\\
\midrule
GPT-2(medium) & 59.12 & 57.52q \\
\midrule
T-5(large) & 58.38 & 56.71 \\
\midrule
FinQANet (BERT-base) & 54.56 & 52.81 \\
\midrule
FinQANet (BERT-large) & 60.67 & 58.99 \\
\midrule
FinQANet (RoBERTa-base) & 64.90 & 63.15 \\
\midrule
FinQANet (RoBERTa-large) & 68.32 & 67.87 \\
\bottomrule
\end{tabular}
}
%\caption{Our main results on test set.}
\caption{Validation results. }
\label{table:valid_res}
\end{center}
\end{table}

\section*{Appendix D: Prompt Details}
For the experiments on GPT-3, here is the list of prompts we used:

\paragraph{Retriever}

Instruction: 
I am a highly intelligent bot. You need to provide me with context and a series of questions. I will respond yes if the context is needed to answer the last question, otherwise, I will respond with no. 

Prompt format: 
context: (supporting fact candidate) questions: (the question sequence up to current question) answer: (yes or no)

\paragraph{Answer-only}
Instruction:
I am a highly intelligent bot. I can have conversations with the user to answer a series of questions. Later questions may depend on previous questions to answer. You need to provide me with the series of questions as the context and I will answer the last question.

Prompt format:
context: (supporting facts) questions: (the question sequence up to current question) answer: (the execution result)

\paragraph{Program-original \& Program-normal}

Instruction:
I am a highly intelligent bot. I can have conversations with the user to answer a series of questions. Later questions may depend on previous questions to answer. You need to provide me with the series of questions as the context and I will answer the last question with a multi-step mathematical solution. We use symbols, such as \#0, \#1, to denote the results of the intermediate steps.

Prompt format:
context: (supporting facts) questions: (the question sequence up to current question) solution: (the program)

\paragraph{CoT Prompt}

Instruction:
I am a highly intelligent bot. I can have conversations with the user to answer a series of questions. Later questions may depend on previous questions to answer. You need to provide me with the series of questions as the context and I will answer the last question with a multi-step mathematical solution with step-by-step explanations. We use symbols, such as \#0, \#1, to denote the results of the intermediate steps.

Prompt format:
context: (supporting facts) questions: (the question sequence up to current question) solution: (CoT explanation and the program).

For all prompts, we add the index of 'Q1', 'Q2', etc., before each question in the question sequence. 

\begin{figure}[htbp]
\centering
\includegraphics[width=0.48\textwidth]{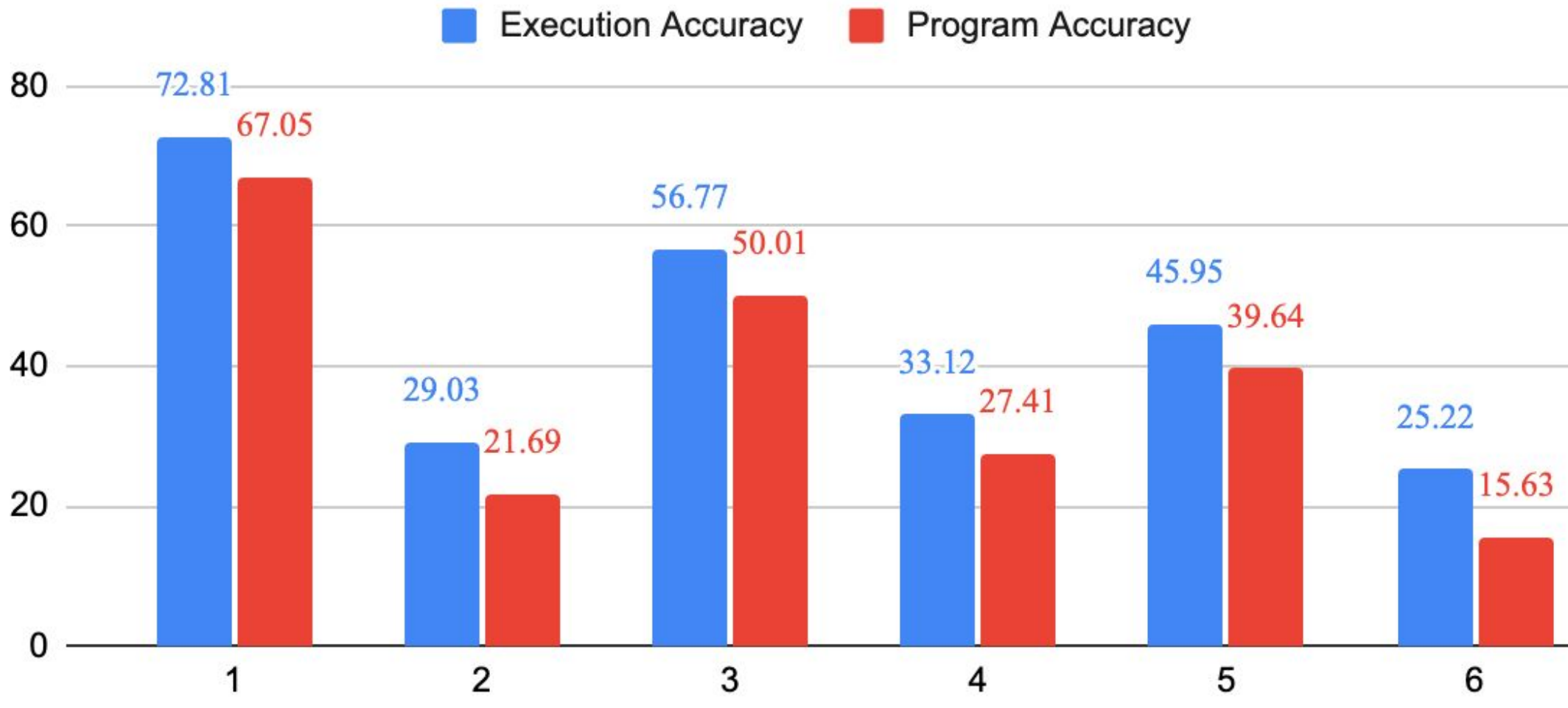}
\caption{Performances for question of the \textit{nth} conversation turn. The second turn mostly makes references to the first turn and GPT-3 often fails to understand it. } 
\label{fig:turn_stat_gpt}
\end{figure}

\begin{figure*}[htbp]
\centering
\includegraphics[width=0.96\textwidth]{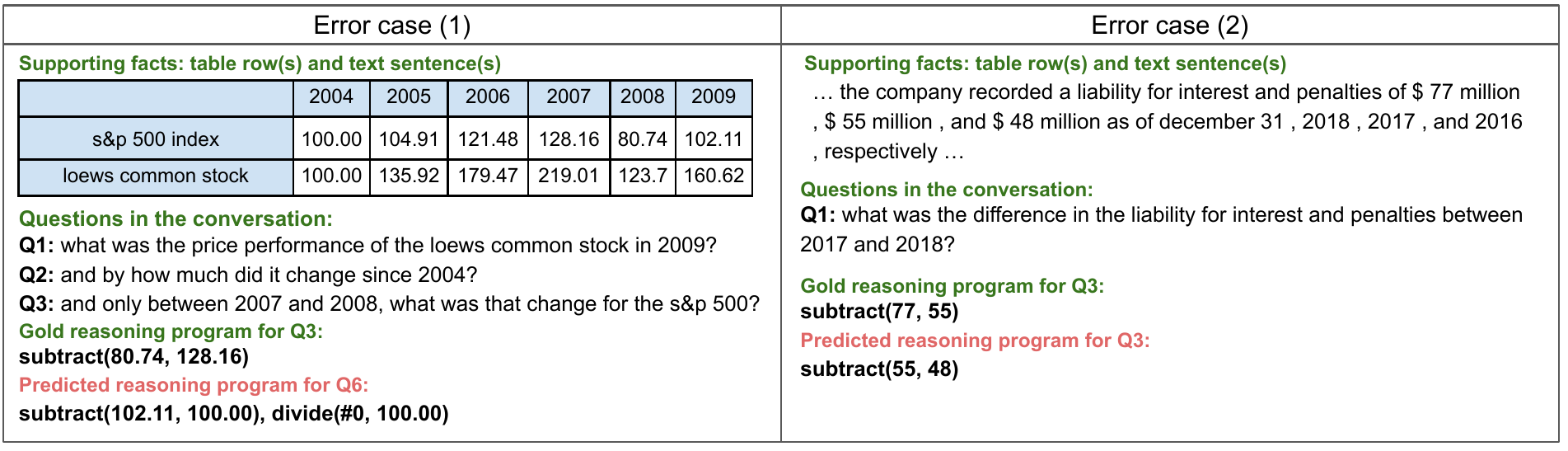}
\caption{Error cases from the results of GPT-3 Program-normal.} 
\label{fig:case_2}
\end{figure*}

Table~\ref{table:gpt_exe_num} shows the results of Program-normal for different number of exemplars. Figure~\ref{fig:turn_stat_gpt} shows the performances for question of the nth conversation turn. Figure~\ref{fig:case_2} gives two error cases of GPT-3 Program-normal. 

\begin{table}[ht]
\small
\begin{center}
\resizebox{.4\textwidth}{!}{%
\begin{tabular}{lcc}
\toprule
\textbf{Exemplar numbers} & \textbf{Exe Acc} & \textbf{Prog Acc}\\
\midrule
5 & 43.52 & 36.62 \\
\midrule
10 & 48.85 & 42.14 \\
\midrule
15 & 49.31 & 44.05 \\
\midrule
20 & 50.30 & 45.10 \\
\midrule
25 & 49.90 & 46.08 \\
\bottomrule
\end{tabular}
}
%\caption{Our main results on test set.}
\caption{Results of Program-normal for different number of exeamplers. }
\label{table:gpt_exe_num}
\end{center}
\end{table}

\end{document}